# Soft-Hard Attention U-Net Model and Benchmark Dataset for Multiscale Image Shadow Removal

Eirini Cholopoulou, Dimitrios E. Diamantis, Dimitra-Christina C. Koutsiou,
Dimitris K. Iakovidis, *Senior Member, IEEE*

*Abstract*—Effective shadow removal is pivotal in enhancing the visual quality of images in various applications, ranging from computer vision to digital photography. During the last decades physics and machine learning -based methodologies have been proposed; however, most of them have limited capacity in capturing complex shadow patterns due to restrictive model assumptions, neglecting the fact that shadows usually appear at different scales. Also, current datasets used for benchmarking shadow removal are composed of a limited number of images with simple scenes containing mainly uniform shadows cast by single objects, whereas only a few of them include both manual shadow annotations and paired shadow-free images. Aiming to address all these limitations in the context of natural scene imaging, including urban environments with complex scenes, the contribution of this study is twofold: a) it proposes a novel deep learning architecture, named Soft-Hard Attention U-net (SHAU), focusing on multiscale shadow removal; b) it provides a novel synthetic dataset, named Multiscale Shadow Removal Dataset (MSRD), containing complex shadow patterns of multiple scales, aiming to serve as a privacy-preserving dataset for a more comprehensive benchmarking of future shadow removal methodologies. Key architectural components of SHAU are the soft and hard attention modules, which along with multiscale feature extraction blocks enable effective shadow removal of different scales and intensities. The results demonstrate the effectiveness of SHAU over the relevant state-of-the-art shadow removal methods across various benchmark datasets, improving the Peak Signal-to-Noise Ratio and Root Mean Square Error for the shadow area by 25.1% and 61.3%, respectively.

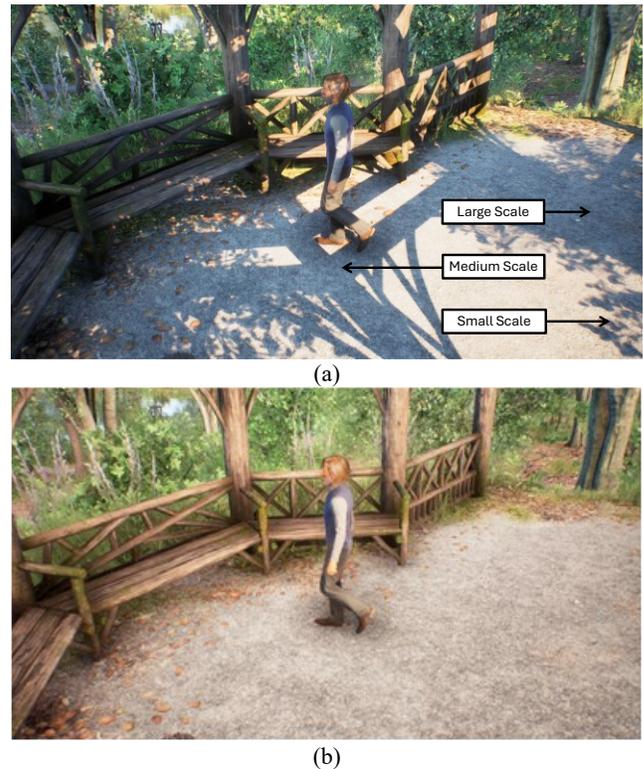

Fig. 1. Representative images from MSRD illustrating the concept of multiscale shadows. (a) Image with shadows. (b) Image without shadows.

## I. INTRODUCTION

SHADOWS are formed as areas of reduced illumination where light is partially or completely occluded by the objects present in a captured scene. Their presence can obscure object boundaries and degrade image quality, diminishing the performance of computer vision algorithms, such as object detection [1].

Shadow removal has been tackled using both physics-based illumination models [2], [3], and machine learning-based [4], [5] models. The physics-based approaches rely on mathematical models with which the illumination/exposure parameters of the images can be estimated to effectively reduce the shadow effects. However, these methods usually require manual interventions, *e.g.*, calculating the intensity of the light reflected at specific image regions, and they have limited capacity in capturing complex shadow patterns due to restrictive model assumptions, *e.g.*, assuming simple lighting conditions and uniform material properties [6]. Machine learning-based shadow removal methods include mainly supervised deep-learning (DL) approaches requiring large, annotated datasets indicating where the shadows are located within the images, or datasets containing pairs of shadowed and shadow-free images. However, collecting and curating such datasets can be time-consuming and costly, and it can be technically challenging or even impractical for some applications, *e.g.* object recognition in outdoor scenes with complex and dynamic lighting conditions. Most of the current datasets developed for benchmarking shadow removal from images contain scenes that are very simple, *e.g.*, including one or a couple of objects, or limited background variation. This can limit the capacity of supervised DL-based shadow removal architectures to adequately model the complexity of shadows in complex scenes resembling those observed in natural environments. Furthermore, a requirement for the assessment of shadow removal is that the benchmarking datasets contain pairs of images with shadows and respective shadow-free images. Therefore, constructing such datasets with natural scenes can be very challenging, which explains the fact that such datasets are currently very limited.

This study focuses on the problem of shadow removal from images with complex scenes containing shadows of multiple scales from different objects (Fig. 1). Aiming to a broader impact, the study focuses on natural scenes, including scenes from urban environments. It proposes a novel supervised DL architecture, named Soft-Hard Attention U-net (SHAU), that follows the U-Net paradigm adapted for multiscale shadow removal. This is achieved by incorporating soft and hard attention modules in its encoder and decoder respectively, along with multiscale blocks connected with residual connections. Aiming to cover the gap related to the lack of sufficiently complex datasets with paired shadow-free images for the assessment of shadow removal, we developed a novel dataset that can serve as a benchmark for this purpose. This dataset, named Multiscale Shadow Removal Dataset (MSRD)[1], consists of 8.5K synthetic images illustrating various scenes in a city park environment. Synthetic data generation has been previously considered to cope with limited data availability in computer vision [7]; however, to the best of our knowledge, MSRD is the first synthetic dataset in the context shadow removal from images.

The rest of this study is organized into four sections. Section II outlines the related work, section III details the proposed methodology, and section IV describes the proposed dataset. The results of a thorough experimental investigation with ablation and comparative studies, are provided in section V. In the last section a discussion and the main conclusions that can be derived from this study are provided along with future research directions.

## II. RELATED WORK

*A. Shadow Removal Methodologies*

Early shadow removal methods primarily leverage prior knowledge to determine the intrinsic physical properties of shadows, such as chromaticity, gradient, illumination [8], as well as, region-specific properties [9]. In [10] a multiscale approach for illumination transfer was proposed in order to recover brightness in the shadow regions. Multiscale image decomposition was used in [11] and [12] to perform shadow removal in individual smaller regions according to their features. Multiscale techniques including shadow scale estimation have been proposed in [13] aiming to restore the brightness in the penumbra region. More recently, a very simple, however effective, fully unsupervised method, called Simple Unsupervised Shadow removal (SUShe), was proposed [14]. That method combines a physics-based optimization algorithm with color feature extraction for shadow detection, and it recovers the luminosity of the shadowed areas by leveraging superpixel segmentation and histogram matching. However, these methods typically struggle to effectively handle diverse shadow patterns, which may lead to the appearance of artifacts along shadow boundaries [15].

Conversely, DL-based approaches have shown promising performance in addressing the limitations of classical methods and producing more accurate shadow removal results. Qu et al. [16] proposed Deshadow-Net, a multi-context feature extraction framework to extract shadow matte. It integrates information from different layers of a Convolutional Neural Network (CNN) to perform shadow removal. The Stacked Conditional Generative Adversarial Network (ST-CGAN) [17] was jointly optimized for both shadow detection and removal using multi-task learning. Le and Samaras [18] employed two deep networks to learn a linear decomposition of the shadow image to facilitate the removal of shadows, which was later extended by incorporating an inpainting network to improve overall performance [19]. Fu et al. [20] formulated shadow removal as a problem of image fusion and employed a neural network to predict the exposure-related parameters. CANet [21] was further proposed as a context-aware framework that leverages the contextual feature information between shadow and non-shadow regions. SG-ShadowNet [22] was proposed as a style-guided shadow removal network that learns the style representation between the shadow and non-shadow regions. CNSNet [23] utilized a "cleanness-guided" strategy for shadow removal that considers the shadow mask for region-wise restoration. Recently, ShadowFormer [24] was proposed as a Retinex-based transformer shadow model which exploits the non-shadow regions to guide shadow region restoration and a multiscale encoder-decoder approach that captures global information [25]. However, CNSNet and ShadowFormer are transformer-based approaches that make use of the multi-head self-attention (MHA) mechanism [26], increasing their computational complexity. Other multiscale and attention-based CNN approaches have been proposed for shadow detection and removal; however, they are domain-dependent addressing shadow removal from remote-sensing images [27], [28].

In addition to the supervised methods, some works focus on addressing the task of shadow removal by leveraging unpaired training data. In this context, MaskShadowGAN [29] used a CycleGAN-based framework to restore the content of shadowed regions from unpaired shadow and shadow-free images, whereas LG-ShadowNet [30] extended this work by incorporating the lightness information to guide the learning process of the network. DC-ShadowNet [31] also used unpaired data to learn a domain-classifier guided network for shadow removal. Despite the promising performance of these methods,

---

[1] This paper has been submitted for possible journal publication. MSRD dataset will be released publicly upon acceptance of the paper to the journal. Currently the dataset is available only upon request to the authors.

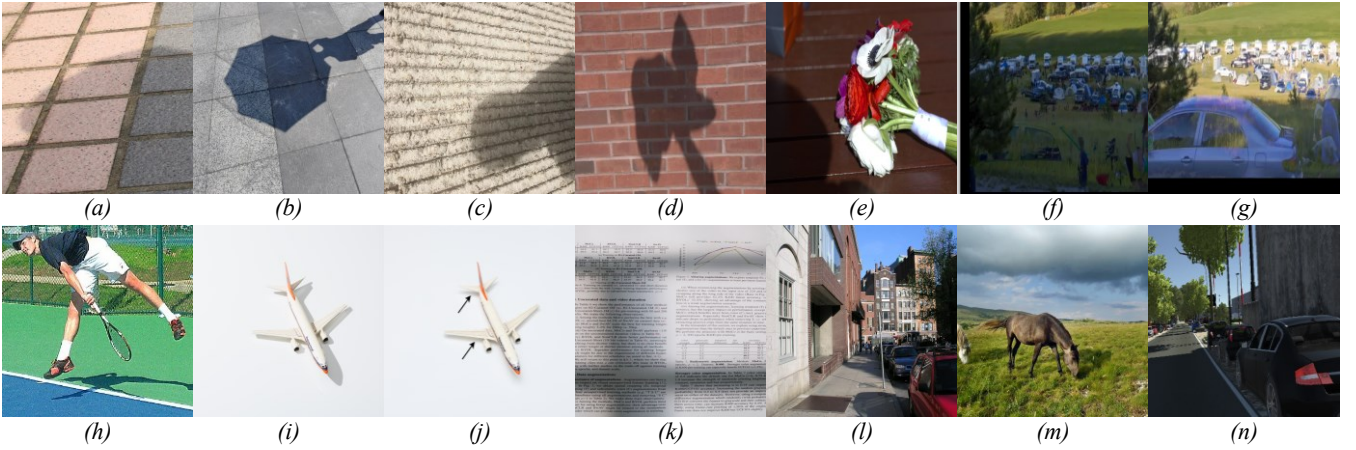

**Fig. 2** Example images of shadow removal datasets that include the (a) SRD, (b) ISTD, (c) USR, (d) UIUC, (e) LRSS, (f) SBU-Timelapse and (g) the respective shadow-free image of (f) with remaining ghost objects, (h) SBU, (i) WSRD shadow-free image and (j) the respective ground truth image of (i) with arrows pointing to remaining shadow patterns, (k) SD7K, (l) CUHK-Shadow, (m) SOBA, (n) Virtual Kitti

TABLE I Shadow Removal Datasets.

| Datasets | Characteristics | | | | | | | |
|---|---|---|---|---|---|---|---|---|
| | Shadow Masks | Shadow-free images | Natural Scenes | Overlapping Shadows | Multiscale Shadows | Synthetic | Number of Images | Image Resolution |
| SRD [16] | - | ✓ | ✓ | - | - | - | 3.088 | 840×640 |
| ISTD [17] / AISTD [19] | ✓ | ✓ | ✓ | - | - | - | 1.870 | 640×480 |
| USR [29] | - | ✓* | ✓ | - | - | - | 2.445 | 450×600, 600×450 |
| UIUC [33] | ✓ | ✓ | ✓ | - | - | - | 108 | 640×425 |
| LRSS [8] | - | ✓** | ✓ | - | - | - | 137 | 5202×3464 |
| SBU [34] | ✓ | - | ✓ | ✓ | ✓ | - | 4.723 | Multiple |
| SBU-Timelapse [19] | - | ✓ | ✓ | ✓ | ✓ | - | 5.649 | 640×480 |
| WSRD [35] | - | ✓ | - | ✓ | - | - | 1.200 | 1920×1440 |
| SD7K [36] | ✓ | ✓ | - | - | - | - | 7.620 | 2462×3699 |
| CUHK-Shadow [37] | ✓ | - | ✓ | ✓ | ✓ | - | 10.500 | Multiple |
| SOBA [38] | - | - | ✓ | ✓ | - | - | 1.000 | Multiple |
| Virtual Kitti [39] | - | - | ✓ | ✓ | ✓ | ✓ | 21.260 | 1242×375 |
| SHAU | ✓ | ✓ | ✓ | ✓ | ✓ | ✓ | 8.582 | 1280×720 |

\* USR contains 1770 unpaired shadow-free images.
\*\* LRSS includes only a subset of 37 images includes shadow-free pairs.

the lack of large-scale annotated paired training data hinders their shadow removal capacity.

*B. Benchmarking Shadow Removal*

Widely used datasets that have been proposed to benchmark the performance of shadow removal methodologies include the Shadow Removal Dataset (SRD) [16], the Image Shadow Triplets Dataset (ISTD) [17] and the Adjusted ISTD (AISTD) [19]. SRD comprises pairs of shadow and shadow-free images, with predicted shadow masks (*i.e.*, not manual annotations) provided by [32]. It mainly illustrates simple outdoor natural scenes with shadows cast by single objects (Fig. 2(a)). The ISTD contains shadow image triplets; each shadow image triplet consists of shadow, shadow-free and shadow mask images, which similarly to SRD, mainly illustrate single object cast shadows of outdoor urban scenes with various backgrounds (Fig. 2(c)). AISTD is a revised version of ISTD with reduced inconsistencies of illumination between the shadow and shadow-free images. A dataset called Unpaired Shadow Removal (USR) [29], offers a collection of shadow and a subset of shadow-free images which are not directly paired, that depict indoor and outdoor environments with most images featuring single object shadows as in Fig. 2(d). The University of Illinois at Urbana Champaign (UIUC) [33] depicts natural scenes where the shadow masks and shadow-free images have been extracted by manipulating the shadows, either by blocking the direct light source or by casting a shadow into the image (Fig. 2 (d)). The Learning to Remove Soft Shadows (LRSS) dataset [8], was proposed for soft shadow removal, consisting of shadow images with only a subset of them having corresponding ground truth images available, depicting outdoor scenes and single objects with uniform background (Fig. 2(e)). Both UIUC and LRSS datasets consist of a relatively small number of samples, that is insufficient for training DL algorithms; thus, increasing the risk of overfitting. A common drawback of all the previously mentioned datasets is that they include mainly simple scenes, with limited variability in objects and backgrounds, and they mostly present well-defined shadow patterns which, in many cases, do not reflect the complexity of real-world shadows. Aiming to increase the complexity of the shadow patterns captured in images, some datasets provide images with overlapping shadows, *e.g.*, in case of objects casting shadows that intersect between each other. These include the SBU [34], SBU-Timelapse [19] and WSRD [35] datasets. The SBU shadow dataset was proposed as a larger dataset with aerial and urban outdoor scenes for benchmarking shadow detection (see Fig. 2(h)); however, it includes only shadow annotations, and it does not include respective shadow-free images; therefore, it is unsuitable in the context of shadow removal. The SBU-Timelapse dataset was later introduced to evaluate shadow removal in timelapse videos containing natural scenes monitored during daytime. This dataset does not contain actual pairs of shadow and shadow-free images. It has been used to extract pseudo pairs of such images by capturing pairs of images with and without shadows of the same scene at different times within a day. However, this way, only approximations of static shadow areas can be obtained, since many scenes include motion shifts and do not remain exactly the same within a day, *e.g.*, in (Fig. 2 (f),(g)). The recently proposed WSRD dataset is intended for assessing shadow removal from objects with various colors and textures, captured under controlled lighting conditions (Fig. 2(i)). Although it contains actual pairs of shadow and shadow-free images, it does not include natural scenes, and in many cases some small-scale shadows can be still noticed within the provided shadow-free images (Fig. 2(j)). Also, it does not contain shadow image triplets, which limits its applicability as a shadow removal benchmark.

Some studies have proposed datasets in the context of document shadow removal, such as the Shadowed Document 7K (SD7K) dataset [36] that contains triplets of shadow, shadow-free and shadow mask images, as illustrated in Fig. 2(k). Various studies have proposed datasets for the task of shadow-detection, such as the Chinese University of Hong Kong (CUHK)-Shadow dataset [37] that contains shadow images with multiple resolutions from three public benchmark datasets for image processing (KITTY, ADE20K, KITTI, USR), images from Google Maps and other web images with various scenes (see Fig. 2(l)) that feature overlapping shadows and the respective hand-annotated shadow masks. The Shadow-Object Association dataset (SOBA) [38] that contains shadow images and images with object annotations of outdoor scenes of multiple resolutions (see Fig. 2(m)). However, since these datasets do not contain shadow-free images, their usage for shadow removal tasks is limited.

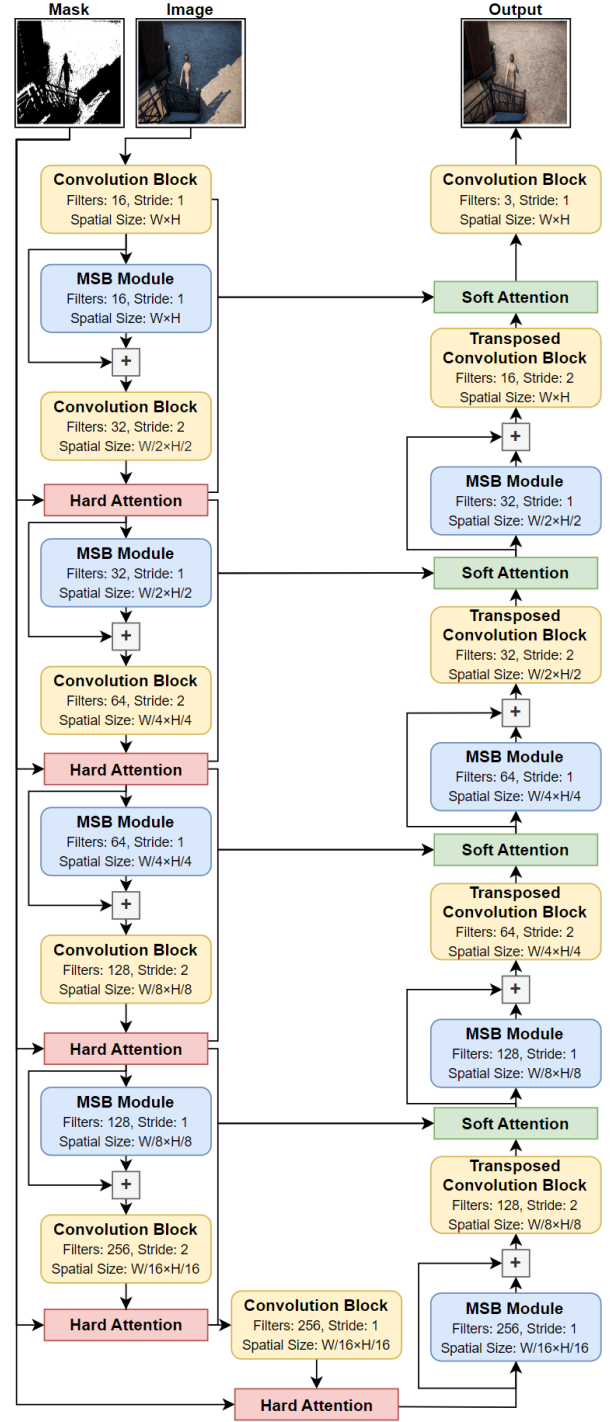

**Fig. 3.** Block diagram of the SHAU architecture.

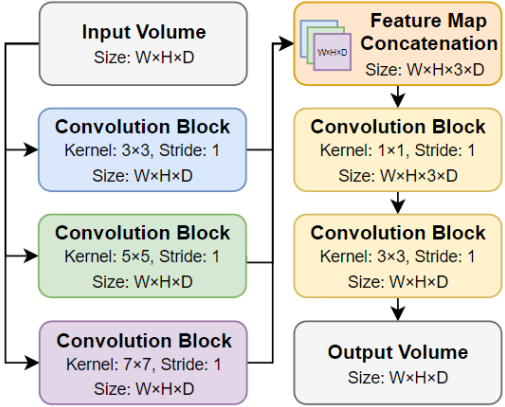

**Fig. 4.** Architecture of the multiscale feature extraction module, where W, H and D represent the width, height and the number of feature maps of the input volume respectively.

Recently, a growing body of research has investigated the use of simulation engines to generate synthetic data that simulate natural scenes with intricate shadow patterns to address computer vision tasks, such as the Virtual KITTI [39] dataset that has been used for shadow detection [40]; however they only present qualitative detection results, since the shadow-free images or shadow masks are not provided (Fig. 2(n)). The characteristics of all the aforementioned datasets are summarized in Table. I.

To the best of our knowledge, the research community lacks a complex large-scale dataset for benchmarking shadow removal from natural scenes, which are characterized by the presence of shadows of multiple scales within the images. This study proposes MSRD, a novel dataset that covers this gap using synthetic images. Advantages of using synthetic images include enhanced determinism in the definition of shadows, flexibility with respect to scene creation and preservation of privacy, as it does not allow for identification of humans or properties. Furthermore, this dataset is sufficiently large for training DL architectures for shadow removal. The following section describes the proposed SHAU architecture, which unlike relevant U-Nets enables the removal of shadows of multiple scales.

## III. SHAU Architecture

The architecture of SHAU extends that of U-Net (Fig. 3) by incorporating a unique combination of multiscale feature extraction blocks, as well as hard and soft attention modules, which enable it to be more effective in shadow removal from images. SHAU receives as inputs: (a) an RGB image with shadows, and (b) a mask of the shadow areas.

Core components of the architecture are the convolutional multiscale feature extraction blocks (MSBs). An illustration of the MSB module is depicted in Fig. 4. The module extracts features under three different scales, *i.e.*, small, medium, and large scale with filter-sizes of 3×3, 5×5 and 7×7, respectively. The extracted features are then concatenated, forming a collection of features. As this collection includes features of different sizes, a pointwise convolution operation [41] is used to aggregate the features by mapping cross-channel correlations [42] between the different scales of detail. The last layer of the MSB module is a 3×3 convolution layer which extracts the same number of features as the input volume. This creates a bottleneck layer, which reduces the number of feature maps extracted from the previous layers and promotes learning [43]. All the convolution layers of the MSB architecture are extracted using a stride size equal to 1, Rectified Linear Unit (ReLU) activations followed by batch normalization.

Choosing to extract features under three different scales enables the module to capture smaller, medium, and larger features from the input space. Using filters larger than 7×7 was deemed impractical as the computational complexity of the model would increase without improvements in generalization performance [44]. Multiscale feature extraction has been used in other architectures, as in the Inception-based models [45]. These architectures typically extract features under two scalesonly, *e.g.*, 3×3 and 5×5. Furthermore, aiming to reduce the number of free-parameters these modules extract a variable number of features per scale. Contrary to this, the MSB module extracts the same number of features across all three scales to not compromise the learning capacity of the module. The free-parameter reduction is instead handled at a later stage, by the bottleneck layer of the module.

The proposed architecture is composed of two primary components: the encoder and the decoder. The encoder is responsible for learning the latent representation of the input image along with its corresponding shadow mask. Subsequently, the decoder uses this latent representation to reconstruct an image that is free of shadows. The encoder of the architecture consists of a single convolution block with stride 1 that extracts 16 feature maps from the input image. This is then followed by four MSB modules, each one having double the number of feature maps from the previous one. The output of MSB module is connected with residual connections, allowing the input to be combined with the output of the module. The spatial dimensionality of each module is then reduced in half by a convolution block with a stride of 2. Employing the convolution block instead of conventional max-pooling, increases the overall non-linearities of the model, promoting learning while simultaneously performing down-sampling [46]. Hard-attention is then used on the down-sampled output in combination with the (soft) image mask of the shadow. Spatial dimensionality reduction of the image mask is achieved using max pooling. The encoder and the decoder are connected using a convolutional block with hard attention.

The architecture of the decoder of SHAU is similar to that of the encoder with the exception of the attention mechanism and the spatial dimensionality reduction. More specifically, the decoder consists of four MSB modules with their output connected to transposed convolution blocks, which up-sample the output volume by two while in parallel reduce the features number in half. Each transposed convolution block is followed by a soft-attention [47] module. This module takes as input the output from the previous block combined with the corresponding output from each MSB module of the same scale from the encoder. The output of the final transposed convolution block, along with the input image, is then used to perform soft attention. To reconstruct the image without shadows, the decoder uses a convolution block with filter size of 3×3, stride of 1 and 3 filters as the output layer.

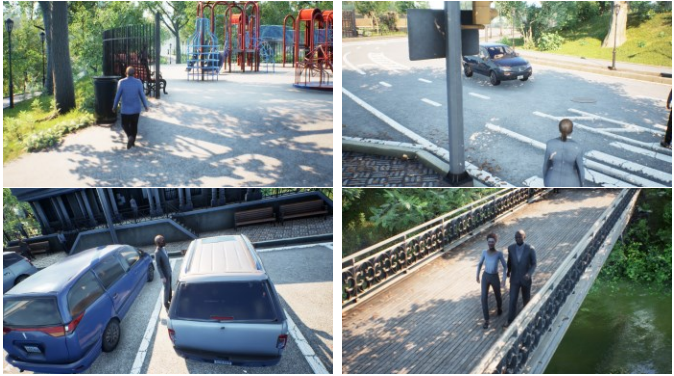

**Fig. 5**. Sample images of the proposed MSRD dataset, depicting various scenes with multi-scale shadows of the virtual environment.

In total, the SHAU architecture is composed of eight MSB modules, with their input and output connected using residual connections. As the architecture is considered relatively deep, while training, it suffers from the problem of vanishing gradient [48]. The residual connections enable the gradient to flow by allowing the input to bypass the intermediate layers of the MSB module, alleviating the vanishing gradient problem. While residual connections that by-pass a single convolution layer have been used in the past by a variety of CNN architectures such as the ones used in ResNet [48], the residual connections used in SHAU by-pass multiple convolution layers. This aims to preserve the extracted features throughout the network and enable deeper MSB modules to capture features from higher levels. Preliminary experiments while designing the architecture validated this approach, demonstrating its effectiveness in enhancing the overall feature extraction capabilities of the network.

Over the years, a variety of U-Net [49] based architectures have emerged. Attention based U-Nets use attention mechanisms to enable the network to better focus on the shadow regions of the image. Following this, SHAU makes use of hard-attention and soft-attention on the encoder and the decoder parts of the architecture respectively. More specifically, on the encoder part of the architecture, the shadow region masks are used to perform spatial attention, guiding the focus of the MSB modules towards input volume regions which include shadows. This enhances the feature representations by prioritizing the shadow region features of the input volume [50], capturing intricate patterns and shadow regions, which are crucial for accurate shadow removal. Through experimentation in the design of the architecture, we found that by using hard spatial attention on the encoder part of the SHAU increases the overall shadow removal capabilities of the model significantly.

Contrary to the hard-attention used in the encoder, the decoder uses soft-attention after every MSB block. This enables the decoder to have fine-grained control over the features maps, enabling dynamic feature localization [51]. While this increases the overall generalization capabilities of the architecture, the decoder is still limited to the features obtained from the hard attention used by the encoder. This results in features that contain only information about the shadow region of the image. To address this limitation, the final convolution layer of the decoder uses soft-attention from features extracted prior to the hard-attention applied on the encoder.

## IV. MSRD Benchmark

### A. Design requirements

The creation of a new dataset for benchmarking shadow removal was motivated by the fact that none of the current datasets proposed for this purpose, fully satisfies all the requirements for a high quality and realistic assessment of shadow removal algorithms. The dataset should provide paired shadow-free images that are devoid of shadow traces and precise shadow masks to facilitate the training and evaluation of shadow removal algorithms. The dataset should cover a variety of scenes that may include urban environments and natural scenes, as well as shots of objects with diverse ranges and angles. This diversity is crucial for creating a challenging environment for training algorithms that can generalize to various scenes and object types. The dataset should also contain shadows of multiple scales, from small specs cast by leaves to large shadows across the ground, in order to be more representative of the shadow patterns found in real-world scenarios (Fig. 1, Fig. 5). These patterns should include overlapping shadows with complex shadow interactions, cast by objects of multiple sizes and diverse distances from the main light source, which is something uncommon in widely used shadow removal datasets (see Fig. 2 (a-e, h, i)). This complexity is required to improve the robustness of shadow removal algorithms in handling multiscale shadows with shadow pattern interactions. Additionally, the dataset should be easily extensible in terms of scenery and object modifications, weather conditions and lighting configuration, with the capacity to augment real-world data for improved performance. This extensibility ensures that the dataset remains relevant for future requirements and relative application objectives. The dataset should also offer high-resolution images that can capture the details of complex environments with fine textures and small objects. A large number of images is also essential to facilitate the training of DL algorithms. Personal privacy should be ensured by avoiding images that contain identifiable individuals or properties, in order to comply with privacy regulations and ensure unlimited access for research purposes without any legal concerns.

### B. Dataset Generation

To create a novel benchmark dataset fulfilling all the requirements listed in the previous paragraph, a fully customizable approach was followed, based on a state-of-the-art rendering software. Specifically, the Unreal Engine (UE) version 5.1, was used for this purpose, which offers a development platform with a high capacity in rendering photo-realistic scenes that accurately simulate real-world lighting conditions and offer fully customizable virtual environments. UE provides a large library of customizable assets, including 3D models, whose appearance is represented by materials that simulate real-world surfaces and define how virtual objects react to lighting. The lighting techniques of UE, include both stationary and dynamic light sources capable of producing various illumination conditions to simulate different times of day. Furthermore, UE offers the *"Metahumans"* tool, which is used for creating highly realistic and customizable human characters, complying with the privacy requirements. These characters are generated as 3D person models, with customizable attributes, such as body type, gender, age, and clothing. To create high-quality image content that can extend beyond 4K resolution with precise control over the scenes, UE offers the *"Sequencer"* tool and cinematic camera actors that can create cinematic sequences that replicate real-world film camera behavior, such as depth of field, focal length and other settings. The cinematic sequences can be exported as images under configurable render settings that include output resolution and format, frame rate and post-processing options.

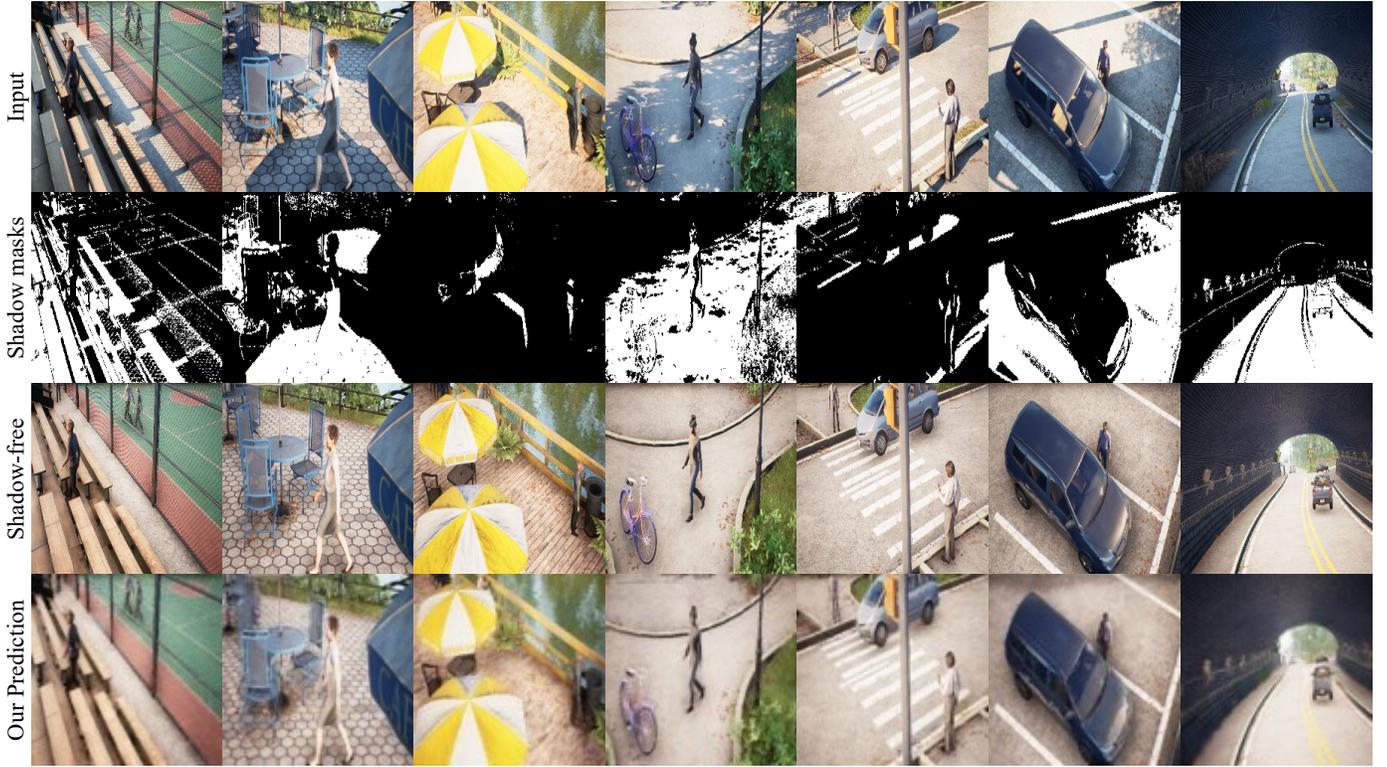

**Fig. 6** Example images from the MSRD benchmark dataset, that include triplets of the input image, the corresponding shadow mask and the ground truth shadow-free image. Results of SHAU are also presented (last row).

The virtual environment considered for the construction of MSRD renders a city park [52] composed of various elements, such as terrain, pedestrians, vehicles, benches, traffic signs, water elements and buildings, as well as a large number of park properties, trees, plants, grass, squares and playgrounds. In the virtual park, these elements can be placed and customized to simulate real-world scenarios, *e.g.*, by altering the position, orientation, and appearance of trees, buildings, and pedestrians. The virtual city park collection includes a total of 524 meshes, 341 image textures, 10 distinct customizable materials and 212 material instances that allow for variations in the material properties for rendering these meshes. These materials were used to produce photo-realistic representations of objects that can be customized by adjusting the material properties of texture, color, glossiness, and reflectivity to replicate different surfaces such as roadways, park benches, trees, and buildings, contributing to the visual realism and diversity of the dataset. The configuration of the intensity and position of the main light source influenced the direction and intensity of shadows, contributing to the complexity and diversity of shadow patterns that can be produced. The pedestrians in the virtual city park environment were created by utilizing the *"Metahumans"* tool.

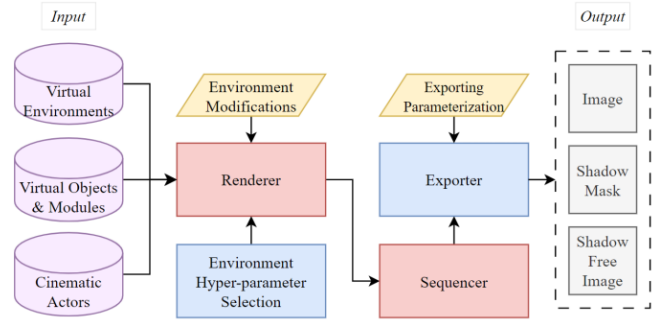

**Fig. 7**. Overall pipeline for the automatic construction of the proposed MSRD dataset.

The randomization of the customizable attributes of gender, body type, age and clothing resulted in the generation of 12 distinct pedestrians with diverse appearances and poses, that were placed throughout the city park. Some examples that showcase the diversity of the virtual city park scenes are presented in Fig. 5 and the first row of Fig. 6.

The proposed MSRD dataset was generated with the use of the cinematic tools of UE, resulting in 8.582 images with a uniform resolution of 1280×720 pixels. The different objects of different sizes included in this environment, such as tree leaves, pedestrians and buildings, enabled casting multiscale shadows in the captured images. All images of MSRD were captured through 10 distinct camera sequences, each generated by a cinematic camera actor placed within the virtual environment. To impose a standard image quality, the selected camera actors shared the same technical characteristics, that are commonly found in real-world cameras. More specifically, each camera had the *"Filmback"* property set to 16:9 Digital Film, which effects the camera's base field of view and aspect ratio, an aperture of 2.8 and a focal length that varied from 20mm to 35mm. The camera sequences were generated by allowing the random movement of the camera actors in the environment under predefined parameters of maximum height and angle ranges. In detail, the cameras were restricted to a maximum height of 3 meters and rotation within a range of 45 to 90 degrees, to provide a diverse range of viewpoints that closely resemble real-world scenarios. Data collection relied on the standard shaders of UE for object rendering and no form of post-processing was applied on the captured images.

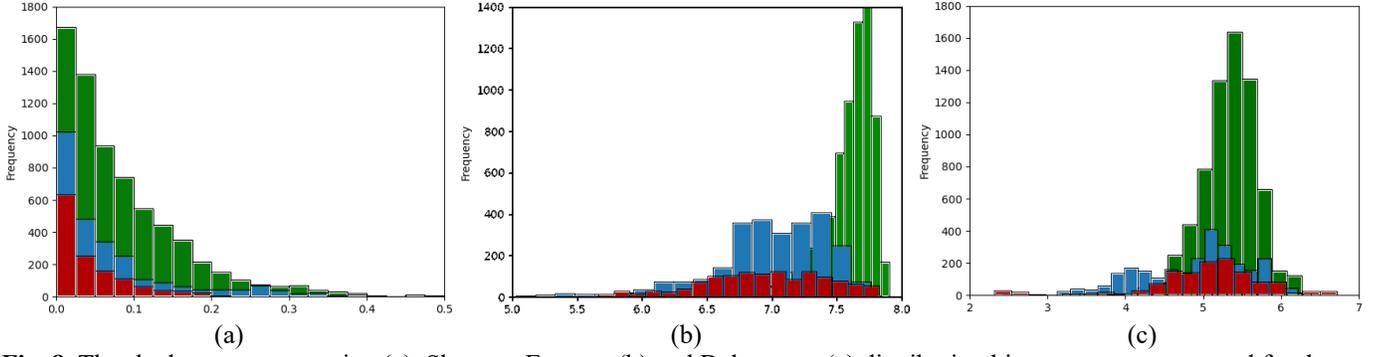

**Fig. 8.** The shadow area proportion (a), Shannon Entropy (b) and Delentropy (c) distribution histograms are presented for the ISTD/AISTD (red), SRD (blue) and MSRD (green) datasets.

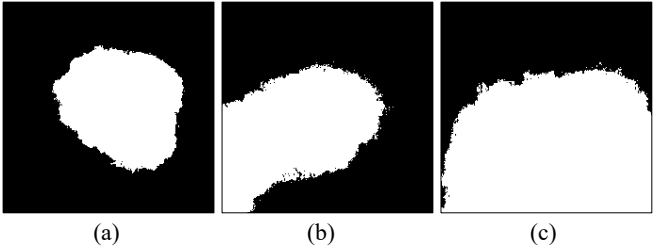

**Fig. 9.** Shadow location distributions of the (a) ISTD/AISTD, (b) SRD and (c) MSRD datasets. White and blue values indicate high and low probability of shadows in that region respectively.

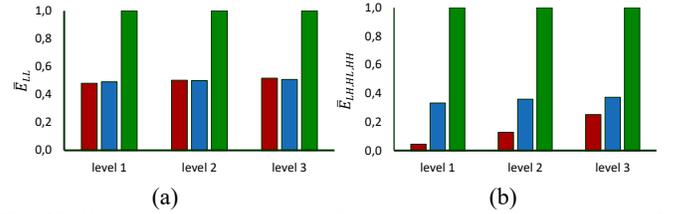

**Fig. 10.** Wavelet analysis results of mean energies at three levels of decomposition for (a) low $\bar{E}_{LL}$ and (b) high $\bar{E}_{LH,HL,HH}$ frequencies for the ISTD/AISTD (red), SRD (blue) and MSRD (green) datasets.

TABLE II Intrinsic dimensionality results and average number of shadows per dataset.

| Dataset | Metric | |
|---|---|---|
| | Intrinsic Dimensionality | Number of shadows |
| ISTD/AISTD | 7.862 | 1.586 |
| SRD | 8.315 | 2.116 |
| **MSRD** | **13.661** | **18.531** |

TABLE III Average entropy results of all datasets.

| Dataset | Metric | |
|---|---|---|
| | Shannon entropy | Delentropy |
| ISTD/AISTD | 6.933 | 4.924 |
| SRD | 6.967 | 4.913 |
| **MSRD** | **7.608** | **5.525** |

The process of generating precise shadow masks involved leveraging the lighting customization options in the environment settings of UE. In the first stage, the camera sequences captured all images under standard lighting settings, including the presence of all cast shadows. To achieve the isolation and subsequent annotation of these shadows, the deactivation of the cast shadow option in UE was selected, which automatically disabled the rendering of all object cast shadows. Then the recapturing of the same scenes followed, this time within the devoid of shadows environment. The generation of the shadow masks was performed by a pixel-wise difference operation between the original images (with shadows) and the recaptured images (without shadows). This process highlighted the precise shadow regions by using a thresholding operation to generate binary shadow masks. The proposed dataset is composed of the original images, shadow-free images, and shadow masks. The overall pipeline for the generation of MSRD is summarized in Fig. 7, which includes the selected virtual environment, additional objects, modules, and cinematic actors, as input to the *'Renderer'* module that renders the virtual scenes by considering user-based environment modifications and hyper-parameter selection, such as lighting and shaders. The rendered scenes are then captured by the *'Sequencer'* tool of UE in order to be exported as images under specific exporting parameterization using the *"Exporter"* part of the pipeline. The *"Exporter"* offers predefined parameterization options, such as image resolution and frame-rate, as well as image processing operations for the extraction of shadow masks, as described in the previous paragraph. This pipeline is extensible and can be used for general-purpose generation of datasets with the use of virtual environments. To the best of our knowledge none of the existing virtual world datasets provide ground truth shadow masks of the generated images, which poses a challenge for benchmarking shadow removal algorithms.

TABLE IV ABLATION STUDY ON SHADOW REMOVAL RESULTS ON THE MSRD DATASET USING DIFFERENT SHAU CONFIGURATIONS.

| Method | PSNR (↑) | | | RMSE (↓) | | |
|---|---|---|---|---|---|---|
| | S | N | A | S | N | A |
| *Baseline* | 28.4 | 34.6 | 27.2 | 6.6 | 5.6 | 6.4 |
| *Attention* | 29.3 | 35.1 | 28.1 | 6.1 | 5.2 | 5.9 |
| *Inception* | 30.1 | 35.6 | 28.2 | 6.0 | 5.1 | 5.7 |
| *MSB* | 31.3 | 35.8 | 28.4 | 5.8 | 4.9 | 5.6 |
| *SHAU* | **32.36** | **36.14** | **28.83** | **5.76** | **4.96** | **5.55** |

TABLE IV SHADOW REMOVAL RESULTS OF SHAU IN COMPARISON WITH STATE-OF-THE-ART METHODOLOGIES ON THE MSRD DATASET.

| Method | PSNR (↑) | | | RMSE (↓) | | |
|---|---|---|---|---|---|---|
| | S | N | A | S | N | A |
| *Input Image* | 15.52 | 24.13 | 14.59 | 40.28 | 16.04 | 31.92 |
| *ST-CGAN* | 18.59 | 16.55 | 14.23 | 25.31 | 26.52 | 25.99 |
| *DC-ShadowNet* | 22.14 | 20.86 | 17.76 | 21.52 | 18.53 | 19.83 |
| *LG-ShadowNet* | 18.63 | 17.49 | 14.24 | 36.86 | 38.56 | 37.82 |
| *SP+M+I-Net* | 16.49 | 27.01 | 16.04 | 30.41 | 11.38 | 23.86 |
| *Fu et al.* | 22.73 | 19.57 | 17.23 | 23.76 | 19.15 | 20.68 |
| *CNSNet* | N/A | N/A | N/A | N/A | N/A | N/A |
| *SG-ShadowNet* | 18.18 | 28.07 | 17.65 | 25.93 | 11.64 | 21.01 |
| *SUShe* | 12.79 | 21.16 | 11.85 | 56.85 | 17.31 | 34.52 |
| *ShadowFormer* | 31.77 | 29.86 | 27.36 | 7.34 | 7.14 | 7.23 |
| *SHAU* | **32.36** | **36.14** | **28.83** | **5.76** | **4.96** | **5.55** |

## V. EXPERIMENTS AND RESULTS

### A. Datasets

The shadow removal performance of the proposed SHAU architecture was evaluated on three benchmark datasets that include ISTD [17] and SRD [16], along with the proposed MSRD dataset. These datasets were selected because they provide triplets of shadow, shadow-free and shadow mask images. The UIUC dataset was not considered as it consists of a relatively small number of samples, that is not sufficient for training DL algorithms [53]. Other datasets were occluded from this study due to limitations such as lack of shadow masks, limited number of images and shadow-free images containing artifacts. ISTD [17] consists of 1330 training and 540 testing triplets, with a resolution of 480×640 pixels in size, that include the shadow images with the corresponding shadow masks and shadow-free images. AISTD [19] was also used as a revised version of ISTD with reduced inconsistencies of illumination between the shadow and shadow-free images. SRD [16] consists of 2.680 training and 408 testing pairs of shadow and shadow-free images of various sizes. Since the respective shadow masks are not provided for SRD, we utilize the predicted masks that are provided by the dual hierarchically aggregation network (DHAN) proposed in [32] for training and testing.

### B. Comparative Analysis of MSRD

To quantitatively compare the proposed MSRD with the current most relevant datasets, the Shannon entropy [54], delentropy [55] and Intrinsic Dimensionality (ID) estimation presented in [56] were considered as measures of dataset complexity. The Shannon entropy scores of the shadow images were computed to quantify the level of variation in pixel intensities, providing insights into the diversity of the image data. High average Shannon entropy values indicate that the dataset contains images with diverse pixel intensity distributions and potentially more challenging patterns for image analysis tasks. The formula for the average Shannon entropy score $H$ of all the grayscale images in a dataset is defined as follows:

$$H = -\frac{1}{N}\sum_{j=1}^{N}\sum_{i=0}^{L-1} p_i log p_i \quad (1)$$

where $N$ is the number of images in the dataset, $L$ is the number of gray levels (256 for 8-bit images), $p_i$ is the probability of occurrence of gray level $i$.

Delentropy was computed to offer insights regarding the underlying spatial image structure. To represent a higher order structure of images, it uses a probability density function, that considers each scalar image pixel value as non-locally related to the entire gradient vector field. This process enables the capture of global image features, by invoking second order properties such as pixel co-occurrence and proximity. For a given image, the calculation of the delentropy measure can be expressed as follows:

$$H(f_x, f_y) = -\sum_{j=1}\sum_{i=1} p_{i,j} log\, p_{i,j} \quad (2)$$

where $p_{i,j}$ is the joint probability density function and $(f_x, f_y)$ are the two derivative components of the gradient field of the given image.

Additionally, the intrinsic dimensionality (ID) of the dataset was assessed to gain insights regarding the geometric complexity of the dataset. Intrinsic dimensionality represents the minimum number of features or dimensions required to capture the underlying structure of a dataset. Given a mapped representation of an image sample $x$, a Maximum Likelihood Estimator (MLE) [57] was employed for the estimation of ID, defined as:

$$m_k(x) = \left[\frac{1}{k-1}\sum_{j=1}^{k-1} log \frac{T_k(x)}{T_j(x)}\right]^{-1} \quad (3)$$

where $k$ is the number of neighbors and $T_k(x)$ is the Euclidean distance from a fixed point $x$ to its $k^{th}$ nearest neighbor. In the context of a given dataset, MLE quantifies complexity by statistically modeling the probability distribution of images. Lower ID scores indicate a dataset with simpler and more uniform structural characteristics, suggesting a reduced degree of complexity, whereas higher ID scores indicate a dataset with a greater degree of structural complexity.

TABLE V RESULTS IN TERMS OF PSNR AND RMSE IN COMPARISON WITH STATE-OF-THE-ART METHODOLOGIES ON THE ISTD, AISTD AND SRD DATASETS.

| Method | Dataset | | | | | | | | | | | | | | | | | |
|---|---|---|---|---|---|---|---|---|---|---|---|---|---|---|---|---|---|---|
| | ISTD | | | | | | AISTD | | | | | | SRD | | | | | |
| | S | | N | | All | | S | | N | | All | | S | | N | | All | |
| | PSNR | RMSE | PSNR | RMSE | PSNR | RMSE | PSNR | RMSE | PSNR | RMSE | PSNR | RMSE | PSNR | RMSE | PSNR | RMSE | PSNR | RMSE |
| *Input Image* | 22.4 | 32.1 | 27.3 | 6.7 | 20.6 | 10.9 | 20.8 | 39.0 | 37.5 | 2.4 | 20.5 | 8.4 | 18.9 | 39.2 | 31.7 | 4.1 | 18.2 | 13.8 |
| *ST-CGAN* | 33.7 | 10.3 | 29.5 | 6.9 | 27.4 | 7.5 | - | 13.4 | - | 7.9 | - | 8.6 | - | 12.6 | - | 6.4 | - | 7.8 |
| *DC-ShadowNet* | 31.7 | 10.6 | 28.9 | 5.8 | 26.4 | 6.6 | 32.0 | 10.3 | 33.5 | 3.5 | 28.8 | 4.6 | 33.4 | 7.7 | 34.9 | 3.4 | 30.5 | 4.7 |
| *LG-ShadowNet* | 31.5 | 11.6 | 29.4 | 5.9 | 26.6 | 6.7 | 32.4 | 10.6 | 33.7 | 4.0 | 29.2 | 5.0 | 28.1 | 20.8 | 31.6 | 4.3 | 25.8 | 8.9 |
| *SP+M+I-Net* | 32.9 | 9.8 | 26.1 | 4.3 | 25.0 | 4.8 | 32.4 | 6.2 | 33.7 | 3.3 | 29.2 | 3.8 | 29.6 | 13.6 | 30.7 | 5.6 | 26.4 | 7.8 |
| *Fu et al.* | 34.7 | 7.8 | 28.6 | 5.6 | 27.2 | 5.9 | 36.0 | 6.5 | 31.2 | 3.8 | 29.4 | 4.2 | 32.3 | 8.6 | 31.9 | 5.7 | 28.4 | 6.5 |
| *CNSNet* | 36.7 | 6.6 | 32.1 | 4.2 | 30.3 | 4.6 | 38.1 | 5.6 | 37.7 | 2.5 | 34.2 | 2.9 | 35.1 | 6.9 | 35.7 | 3.3 | 31.7 | 4.3 |
| *SG-ShadowNet* | 31.3 | 10.5 | 31.2 | 3.9 | 27.6 | 4.9 | 32.6 | 8.7 | 27.6 | 3.3 | 26.2 | 4.0 | 28.7 | 10.7 | 29.2 | 3.8 | 25.5 | 5.7 |
| *SUShe* | 30.1 | 14.9 | 27.3 | 6.7 | 24.7 | 8.1 | 26.4 | 12.2 | 37.4 | 2.5 | 30.1 | 4.1 | 24.6 | 20.2 | 28.7 | 5.7 | 22.6 | 9.7 |
| *ShadowFormer* | 38.2 | 6.0 | 34.3 | 3.7 | 32.2 | 4.1 | 39.7 | 5.2 | 38.8 | 2.3 | 35.5 | 2.8 | 36.9 | 5.9 | 36.2 | 3.4 | 32.9 | 4.0 |
| *SHAU* | **39.1** | **5.9** | **34.4** | **3.3** | **33.4** | **3.9** | **40.1** | **5.0** | **40.4** | **1.8** | **35.6** | **2.5** | **37.1** | **5.5** | **38.1** | **2.4** | **33.4** | **3.2** |

The results of the average Shannon entropy and delentropy for the MSRD, ISTD and SRD datasets are reported in Table II. It can be observed that the highest scores are attributed to the MSRD dataset for both Shannon entropy and delentropy, estimated at 7.61 and 5.52, respectively. This indicates a higher degree of diversity and information content on both local and global scale, respectively. The Shannon entropy and delentropy distributions for all datasets are presented in Fig. 8(b) and (c), highlighting the variability and information content within each dataset.

Table III presents the ID scores for all datasets. Reportedly, the proposed dataset achieves the highest score of 13.66, indicating greater degree of structural complexity and pattern interaction. In contrast, ISTD and SRD are characterized by lower scores of 7.86 and 8.31, respectively. These scores can be attributed to the uniform object patterns present in the images of the respective datasets.

To investigate the degree to which the compared datasets include shadows of multiple scales, we have performed wavelet analysis, by applying the 2D Discrete Wavelet Transform (DWT) on the shadow masks of the compared datasets. Multi-level DWT operates at different levels of image decomposition, each capturing representations of progressively finer detail at low (LL) and high frequencies in the horizontal (LH), vertical (HL) and diagonal (HH) direction. The resulting low and high-frequency approximation coefficients at each level were used to calculate the energy values, to evaluate the complexity of shadow patterns within the datasets, where high energy values indicate rich information content with high variability in images. The mean energy values at three levels of DWT decomposition are presented in Fig. 10, normalized by the maximum value of each respective frequency coefficient. It can be observed that for both low (Fig. 10 (a)) and high (Fig. 10 (b)) frequencies, the proposed MSRD dataset has the highest mean energies $\bar{E}_{LL}$ and $\bar{E}_{LH,HL,HH}$ in comparison with ISTD and SRD datasets, indicating higher information content at multiple levels of detail.

To further validate the compliance of the proposed dataset with the described design requirements, we computed the statistical complexity measures presented in [37]. These measures include the estimation of the shadow area proportion of pixels occupied by shadows in each image, the number of shadows per image and the shadow location distribution per dataset. The histogram plots of the shadow area proportion for each dataset are displayed in Fig. 8(a). It can be observed that most images in ISTD and SRD are characterized by smaller shadow regions, whereas MSRD contains more diverse shadow patterns. Furthermore, the shadow location distributions that indicate the per-pixel probability of shadow occurrence are presented for all datasets in Fig. 9. Notably, the shadow patterns of the MSRD dataset cover a wider spatial area, which can indicate greater probability of overlapping shadows, excluding the upper region often associated with the sky. In contrast, ISTD and SRD display shadows that are primarily concentrated in the center region of the image. Table III reports the average number of shadows per image for all datasets, where it can be observed that ISTD and SRD have approximately 1.59 and 2.12 shadow regions per image, while MSRD contains on average 18.53 shadow regions per image, validating the appearance of multiscale shadows in the images.

*B. Shadow Removal Evaluation*

The evaluation of the performance of the proposed SHAU architecture in removing shadows from images at multiple scales, is evaluated both qualitatively and quantitatively in the following sections.

**Experimental Setup.** The comparative evaluation of the proposed SHAU architecture involved nine state-of-the-art shadow removal methods that cover a diverse range of approaches that rely on different types of networks to remove shadows, including ST-CGAN [17], DC-ShadowNet [31], LG-ShadowNet [58], SP+M+I-Net [19], Fu et al. [59], CNSNet [23], SG-ShadowNet [22], ShadowFormer [24] and a physics-based method, SUShe [14], which does not require training. For a fair comparison, the results are reproduced utilizing the official source code and hyperparameters of each reported method, with the exception of CNSNet for which the source code was not available.

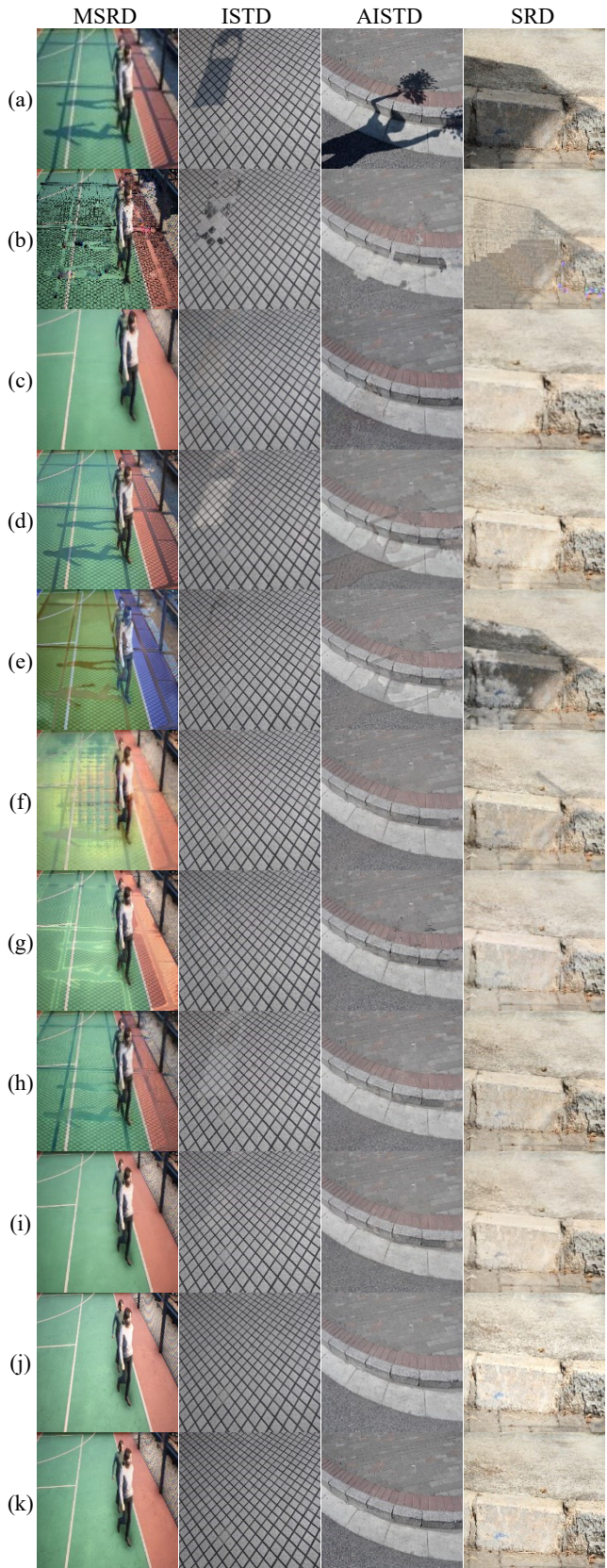

**Fig. 11** Qualitative results among SHAU and state-of-the-art shadow removal methods that include the (a) input image, (b) SUShe, (c) ST-CGAN, (d) DC-ShadowNet, (e) LG-ShadowNet, (f) SP+M+I-Net, (g) Fu et al., (h) SG-ShadowNet, (i) ShadowFormer, (j) SHAU, and (k) the ground truth.

The proposed methodology is implemented in Python 3.8 using the TensorFlow 2.11 framework. The number of training epochs was set to 2,000. The Adam [60] optimizer was used with the learning rate set to 0.0001. The resolution of the input images was set to 256×256 pixels in size and data augmentation included random image scaling and rotations.

To quantify the performance of the proposed methodology, we utilized the Peak Signal-to-Noise Ratio (PSNR) [61] and the Root Mean Square Error (RMSE) as widely used evaluation metrics assessing the quality of the output images. As adopted by previous works [16], [18], [22], [32], [59], [62], we compute the RMSE between the produced shadow-free and the ground-truth images in the LAB color space. PSNR is computed in the RGB color space. Higher values of PSNR and lower values of RMSE indicate better performance respectively.

**Ablation Study.** To better understand the effects of each component of the proposed SHAU architecture, we conducted a series of ablation studies. We implemented four variations of the SHAU architecture and evaluated each one on the proposed MSRD dataset. On the first variation, which is considered as the baseline ($SHAU_{Baseline}$), we removed the multiscale module along with the soft-attention and the residual connections. For fair comparison, the last soft-attention used in the SHAU architecture in the baseline version was replaced with a residual connection enabling the decoder to have high level features of the original image. The results in Table IV outline that the baseline model performs significantly lower in the shadow regions (~13%) when compared with the SHAU architecture, in both PSNR and RMSE metrics. The second and third versions tested the effectiveness of the multiscale module of the architecture. More specifically, on the second version we replaced the MSB modules of the architecture with Inception modules [45] in ($SHAU_{Inception}$) and on the third we used the proposed MSB modules ($SHAU_{MSB}$). In both cases, the multiscale module increased the overall shadow removal performance of the baseline architecture, yet as it can be observed in the results, the proposed MSB version of the architecture outperforms the Inception-based version in both PSNR and RMSE metrics across all regions of the image. This confirms that the multiscale feature extraction component of the architecture enables the model to better capture multiscale shadows appearing in the dataset. The last version ($SHAU_{Attention}$) examined the role of soft-attention on the decoder level of the architecture. The results showed that, although the contribution of the soft-attention mechanism is noticeable, it does not contribute equally to the multiscale feature extraction component of the architecture.

**Quantitative Results.** Table. V. demonstrates the comparative results obtained on the proposed MSRD dataset among all methodologies. It can be observed that the proposed SHAU architecture outperforms the state-of-the-art methodologies on all benchmark datasets including the more complex MSRD dataset. Although the performance of SHAU is higher on both shadow and non-shadow regions of the image when compared with the state-of-the-art, on the non-shadow regions the generalization capabilities of the model are more evident. This is important, as the architecture is able to effectively remove the shadows from the input image while maintaining the characteristics of the non-shadow regions.

The results obtained from the comparative experiments on the benchmark ISTD, AISTD and SRD datasets are demonstrated in Tab. VI. in terms of RMSE and PSNR. It can be observed that SHAU outperforms all competing methods, for the shadow area (S), non-shadow area (N) and entire image area (A), in all datasets. Notably, most state-of-the-art methods [24], [59], [62] require the pre-computation of dataset specific regularization parameters, that typically use least squares methods and mask erosion or dilation [19], to facilitate the training process. Conversely, SHAU does not rely on generating such parameters beforehand and still manages to effectively capture the intricate patterns of shadowed regions for their effective removal.

**Qualitative Results**. To further demonstrate the capacity of the proposed SHAU architecture against state-of-the-art methods, a visual illustration of the shadow removal evaluation is presented in Fig. 11., which presents the original input images, the ground truth images along with the predicted shadow-free images obtained among all methodologies for all datasets. In the predicted samples of MSRD presented in Fig. 11, methods such as SG-ShadowNet, DC-ShadowNet and LG-ShadowNet preserve large parts of the shadow areas in their predictions, while others including SP+M+I-Net and Fu. *et al*. restore the shadow regions with inconsistencies that result in shadow boundaries and surface structure artifacts. In the case of the ISTD, AISTD and SRD datasets, most current methods leave ghost objects in the predicted shadow-free images, that result in sharp shadow boundaries, as shown in Fig. 11(c) for ISTD, Fig. 11(d) for AISTD and Fig. 11(f) for SRD. In contrast, the proposed SHAU architecture successfully removes the shadow regions and effectively restores these areas without leaving shadow traces, boundary artifacts or illumination inconsistencies, as well as it effectively removes shadows at multiple scales that other methods capture only partially. Additional examples of the predicted results of the proposed SHAU architecture are illustrated in Fig. 6, along with the corresponding shadow masks and shadow-free images for qualitative evaluation.

## VI. CONCLUSIONS

This study presented a novel DL architecture named Soft-Hard Attention U-net (SHAU) which incorporates hard and soft attention modules in combination with multiscale feature extraction blocks for multiscale shadow removal. Furthermore, considering the limitations of the existing shadow removal benchmark datasets, we presented a novel synthetic Multiscale Shadow Removal Dataset (MSRD), which contains high-resolution images depicting natural scenes with complex shadow patterns of multiple scales. The evaluation of this work is twofold: a) MSRD was assessed in comparison to most relevant current shadow removal datasets (ISTD, AISTD and SRD), using statistical measures such as entropy, intrinsic dimensionality and DWT analysis; b) the performance of the SHAU architecture was assessed by conducting comparative experiments between nine state-of-the-art shadow removal methods, in terms of RMSE and PSNR metrics on the ISTD, AISTD and SRD datasets and the proposed MSRD.

Overall, the following conclusions can be derived:
- SHAU improves the PSNR and RMSE performance metrics for the shadow image area by 25.1% and 61.3%, respectively, among all datasets over the state-of-the-art methods.
- The results indicate that SHAU is able to remove shadows in complex environments which simulate real situations.
- To the best of our knowledge MSRD is the largest and most comprehensive dataset for shadow removal in complex scenes.
- The obtained results demonstrated the higher complexity of MSRD in comparison to other benchmark shadow removal datasets.
- MSRD offers an extensible framework for dataset generation using customizable virtual environments, with controlled lighting conditions and configurable image capturing settings.

Although the proposed SHAU architecture is capable of effectively removing complex shadows from images, it requires shadow masks to be available. While this limitation can be addressed by employing a pre-processing shadow detection module, in future work we aim to enhance the architecture to perform shadow detection and removal in an end-to-end manner, as well as by investigating the use of domain adaptation techniques for synthetic-to-real image translation.

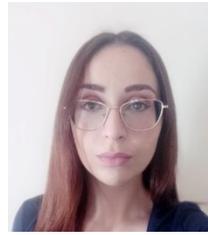

**Eirini Cholopoulou,** is a Ph.D. candidate at the department of Computer Science and Biomedical Informatics, School of Science, University of Thessaly from which she received her B.Sc. in Computer Science and Biomedical Informatics. The scope of her doctoral research covers Machine Learning-based Modeling and Explainable Decision Making. Her research interests include computer vision, anomaly detection and interpretable Artificial Intelligence (AI).

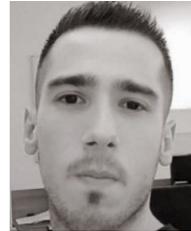

**Dimitrios E. Diamantis,** was born in Athens, Greece, in 1991. He received the B.Sc. degree in computer science from the Department of Computer Engineering, Technological Educational Institute of Central Greece, in 2014, and the M.Sc. degree (Hons.) in computational medicine and biology and the Ph.D. degree in deep learning from the University of Thessaly, Greece, in 2017 and 2021, respectively. From 2009 to 2024, he was a Senior Software Engineer, a Solution Architect, and a Chief Technical Officer (CTO) in several worldwide enterprises, where he gained significant experience in the enterprise software industry. Since 2016, he has been a member of the Biomedical Imaging Laboratory, University of Thessaly. His research interests include signal, image and video analysis, intelligent systems, deep learning, software engineering, and applications.

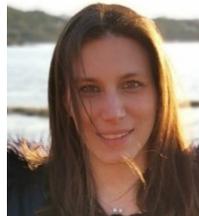

**Dimitra-Christina C. Koutsiou,** received a BSc degree in Physics and a MSc degree in Astrophysics, Astronomy and Mechanics, from the University of Athens and a PhD in Computer Vision from the University of Thessaly. Her research interests focus on image processing applications such as color image segmentation, shadow detection and removal etc. using machine learning algorithms. She has participated in research projects co-funded by the E.U. and the Greek national funds and her PhD research was funded by the State Scholarships Foundation.

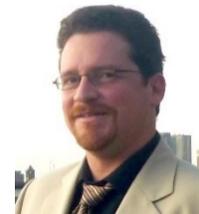

**Dimitris K. Iakovidis,** (Senior Member IEEE) received his PhD in Informatics in 2004 from the University of Athens in Greece. He holds a Full Professor position at the Department of Computer Science and Biomedical Informatics of the University of Thessaly in Greece, and he is the Director of the Biomedical Imaging Laboratory of that department. His research interests include signal/image processing, decision support systems, intelligent systems, and applications. In this context he has co-authored over 200 papers in international journals, conferences, and books. Dr. Iakovidis has served as Associate Editor of IEEE Transactions on Fuzzy Systems.